\documentclass[lettersize,journal]{IEEEtran}
\usepackage{amsmath,amsfonts}
\usepackage{algorithmic}
\usepackage{algorithm}
\usepackage{array}
\usepackage[caption=false,font=normalsize,labelfont=sf,textfont=sf]{subfig}
\usepackage{textcomp}
\usepackage{stfloats}
\usepackage{url}
\usepackage{verbatim}
\usepackage{graphicx}
\usepackage{booktabs}
\usepackage{cite}
\usepackage{multirow}
\begin{document}

\title{Multi-Dimensional Refinement Graph Convolutional Network with Robust Decouple Loss for Fine-Grained Skeleton-Based Action Recognition}

\author{Sheng-Lan Liu, Yu-Ning Ding, Jin-Rong Zhang, Kai-Yuan Liu, Si-Fan Zhang, Fei-Long Wang, and Gao Huang
        % <-this % stops a space
\thanks{Sheng-Lan Liu, Yu-Ning Ding, Jin-Rong Zhang, Kai-Yuan Liu, Si-Fan Zhang, and Fei-Long Wang are with the Computer Science and Technology, Dalian University of Technology, Dalian 116024, China. E-mail: (liusl@dlut.edu.cn; \{rookie233, zjr15272565639, 1154864382, 201981131\} @mail.dlut.edu.cn; wangfeilong@dlut.edu.cn)}% <-this % stops a space
\thanks{Gao Huang is with Department of Automation, Tsinghua University, Beijing, China. Email: gaohuang@tsinghua.edu.cn.}
\thanks{(Shenglan Liu and Yuning Ding contributed equally to this work.)}
\thanks{(Corresponding author: Sheng-Lan Liu.)}}

\markboth{Journal of \LaTeX\ Class Files,~Vol.~14, No.~8, August~2021}%
{Shell \MakeLowercase{\textit{et al.}}: A Sample Article Using IEEEtran.cls for IEEE Journals}

\IEEEpubid{}

\maketitle

\begin{abstract}
Graph convolutional networks have been widely used in skeleton-based action recognition. However, existing approaches are limited in fine-grained action recognition due to the similarity of inter-class data. Moreover, the noisy data from pose extraction increases the challenge of fine-grained recognition. In this work, we propose a flexible attention block called Channel-Variable Spatial-Temporal Attention (CVSTA) to enhance the discriminative power of spatial-temporal joints and obtain a more compact intra-class feature distribution. Based on CVSTA, we construct a Multi-Dimensional Refinement Graph Convolutional Network (MDR-GCN), which can improve the discrimination among channel-, joint- and frame-level features for fine-grained actions. Furthermore, we propose a Robust Decouple Loss (RDL), which significantly boosts the effect of the CVSTA and reduces the impact of noise. The proposed method combining MDR-GCN with RDL outperforms the known state-of-the-art skeleton-based approaches on fine-grained datasets, FineGym99 and FSD-10, and also on the coarse dataset NTU-RGB+D X-view version.  

\end{abstract}

\begin{IEEEkeywords}
Graph convolutional network, fine-grained action, Robust Decouple loss, Spatial-Temporal attention
\end{IEEEkeywords}

\section{Introduction}
\label{sec:intro}

Skeleton-based action recognition has been an attractive emerging topic because of its excellent robustness in dynamic environments and human-centered applications. In recent years, fine-grained action tasks are followed with interest in many fields. However, the fine-grained action recognition task remains challenging due to new difficulties.

\begin{figure}[t]
\centering
\includegraphics[width=1\columnwidth]{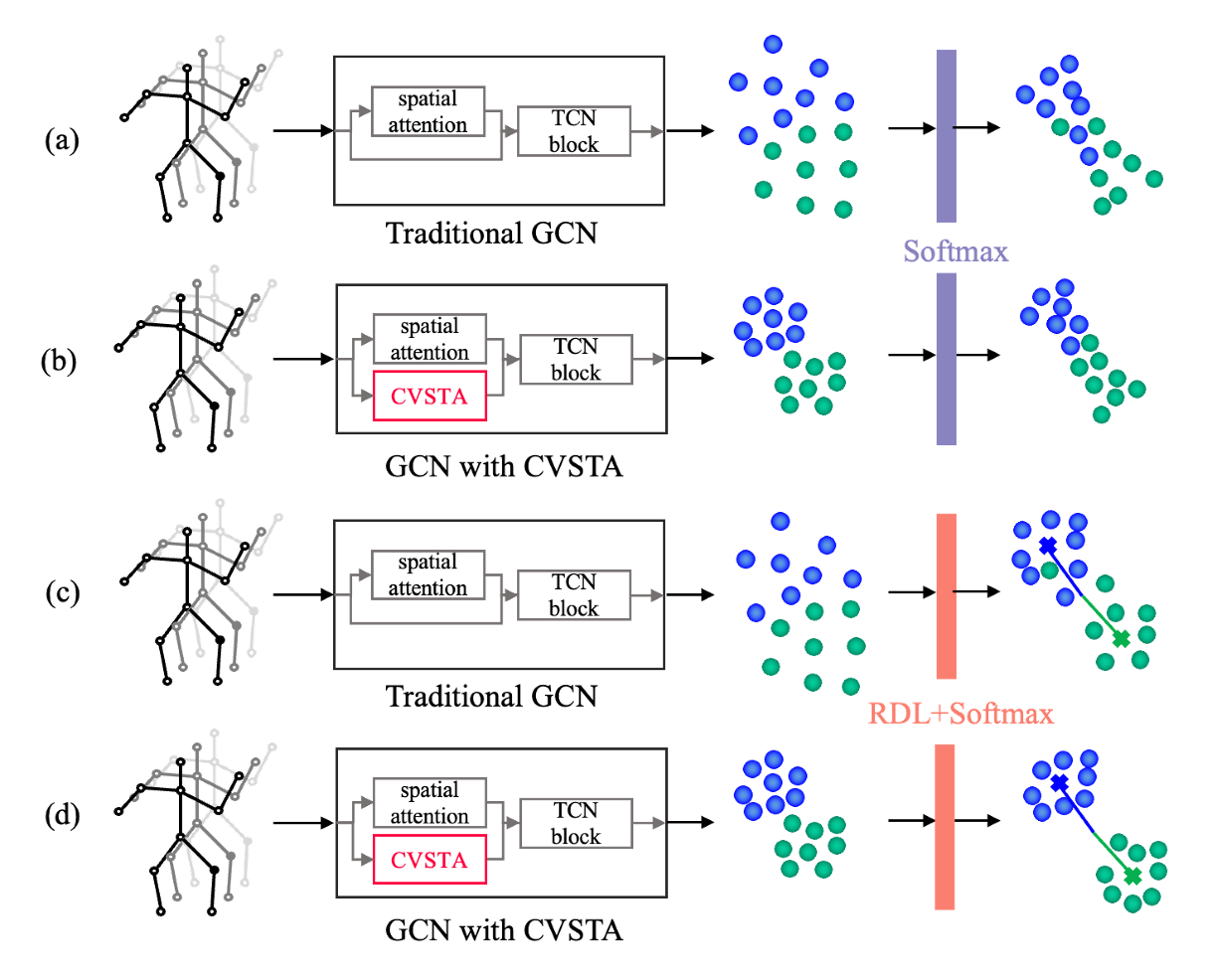} % Reduce the figure size so that it is slightly narrower than the column. Don't use precise values for figure width. This setup will avoid overfull boxes.
\caption{Schematic diagram of feature distribution under method (a) traditional GCN with softmax loss (b) GCN+CVSTA with softmax loss (c) traditional GCN with RDL+softmax (d) GCN+CVSTA with RDL+softmax. The input is the skeleton sequences for two types of actions. The blue and green points represent the feature distribution of different actions.}
\label{fig0}
\end{figure}

The first challenge is to explore the \emph{discriminative power of spatial-temporal joints}. This indicates that inter-joint and inter-frame relationships vary with different actions and their types \cite{zhong2022spatio}. Early RNN or CNN methods \cite{du2015hierarchical, liu2016spatio, song2017end, liu2017two, hou2016skeleton} usually model the skeleton data as a sequence of the coordinate vectors or a pseudo-image but ignore the dependencies between joints. Many frontier GCN-based methods \cite{STGCN, 2SAGCN, song2020richly, gao2021skeleton, CTRGCN, shu2022expansion, chi2022infogcn} apply graph convolution with fixed and learnable parameter matrices to achieve spatial pattern modeling. However, the previous GCN-based methods lack considering the advantages of both channels and spatial-temporal attention. For multi-modality data, STAR-transformer \cite{ahn2023star} adopts the method of adding spatial-temporal fusion attention after feature extraction, but for skeletal data, this approach will cause feature redundancy and lead to performance limitations. Efficient-GCN \cite{Effi} has concentrated on spatial-temporal relationships by modeling a new attention block in fixed channels, which causes a lack of adaptive representation capability of channels and the difficulty of further expansion on the variable channel models. Therefore, the spatial-temporal discriminative capability in the existing models is insufficient for the fine-grained recognition task. 

The second challenge is to obtain \emph{robust and discriminative embedding sample distributions} of skeleton-based actions. For fine-grained recognition tasks, both separable and discriminative learned features should be considered in loss functions\cite{chang2020devil}. In addition, inevitable outliers and noises also should be noticed. Some existing work \cite{li2021trustable, li2022selective} offer solutions for noise labels, but the outliers in collected data of joints are still difficult to deal with. The conventional softmax loss mainly encourages the separability of features, which causes weak intra-class compactness even if the model is more advanced. The explicit methods \cite{wen2016discriminative, he2018triplet} achieve maximized inter-class and minimize intra-class variance (one or both) by utilizing an additional loss. However, existing explicit methods are weak-robust for outliers and numerically unbalanced with softmax loss. Angular softmax approaches \cite{liu2016large, liu2017sphereface, wang2018additive} solve the problems by normalizing features and class centers in softmax. Some scale and margin versions of angular softmax are also proposed to enhance the discriminative capability of features \cite{deng2019arcface, kim2022adaface}. These implicit methods are challenging to optimize norms and margins for each class adaptively (the details will be presented in Sec 2.2 and 3.3). Considering the deficiencies in existing loss functions, we need a more effective loss to achieve more robust and discriminative embedding for the fine-grained recognition task.

In this paper, we propose a dynamic spatial-temporal attention block called Channel-Variable Spatial-Temporal Attention (CVSTA) to solve the problems above. It can build a comprehensive connection between frames and joints and capture more potential features. Besides, as a flexible block, it could be easily combined with other GCNs to get better performance. In contrast to traditional GCN-based methods, our approach obtains spatial-temporal features more sufficiently, which promotes the clustering of intra-class samples. Therefore, to enhance the discrimination among channel-, joint- and frame-level, we propose a Multi-Dimensional Refinement Graph Convolution Network (MDR-GCN), which includes the channel-wise GCN framework with CVSTA and enhanced temporal convolution blocks.

To fully consider the noises and outliers of skeleton data, motivated by related fine-grained tasks \cite{zhang2019cross, peeples2022learnable}, we propose a Robust Decouple Loss (RDL). RDL achieves large class-wise discriminative embedding by leveraging the ratio of norms of each class along with both intra-class and inter-class cosine of the vectors, which significantly improves the robustness of the existing explicit loss function. Thus, RDL can optimize variances along the cosine/norm aspect(s) and reduce the numerical discrepancy of losses, further facilitating the role of CVSTA. Besides, by decoupling center loss \cite{wen2016discriminative}, RDL has more scales that can adjust the impact of noisy data. Figure \ref{fig0} shows the improvement in the feature distribution arising from RDL and CVSTA.

Our contributions are summarized as follows:

1. We propose a new GCN model named MDR-GCN for fine-grained action recognition by utilizing our designed CVSTA block, which captures relationships of joints across frames in variable channels. 

2. We propose RDL by decoupling the center loss for robust discriminative embedding. In our experiments, RDL enhances the performance of CVSTA and outperforms state-of-the-art losses.

3. The proposed method combining MDR-GCN with RDL outperforms the known state-of-the-art skeleton-based approaches on the FineGym99, the FSD-10, and the NTU-RGB+D datasets. 

\section{Related Work}
\label{sec:rwork}

\subsection{Skeleton-based action recognition} 

Early skeleton-based action recognition methods mainly employ RNN or CNN to extract discriminative features. RNN-based approaches \cite{du2015hierarchical, song2017end, li1805skeleton, jiang2019action} generally explain action features as multi-dimensional time series, focusing on extracting actions' temporal features rather than exploiting spatial ones. To deeply realize spatial characteristics, CNN-based networks \cite{henaff2015deep, liu2017two, soo2017interpretable, li2017skeleton, cao2018skeleton, li2021memory} transform the skeleton data into grid images to simplify the training process. However, neither of the two approaches can model the structured dependencies of skeletons because of the inherent calculation strategy.

In recent years, researchers have become increasingly interested in GCN-based methods \cite{shuman2013emerging, kipf2018neural, ye2020dynamic, cheng2020skeleton, qin2022fusing} which can reflect the structured relationships between skeletons. Most studies have focused on spatial modeling, which contains pre-defined \cite{STGCN}, learnable \cite{2SAGCN}, and dynamic \cite{CTRGCN} ways. ST-GCN \cite{STGCN} utilizes the original heuristically pre-defined graph physically driven by the human body, which hardly realizes the dependencies between unlinked joints. As a learnable method, 2S-AGCN \cite{2SAGCN} is further proposed to capture data-driven graphs for more dependencies of skeletons in shared channels. To explore more types of motion features in channel view, some researchers suggested dynamic independent channel-based models \cite{DCGCN, CTRGCN} by offering more graph topologies along with channels. DC-GCN \cite{DCGCN} sets individually parameterized topologies for different channel groups, but it is hard to optimize because of excessive parameters. Integrating the shared and learnable channel-wise topologies is a practical scheme. CTR-GCN \cite{CTRGCN} leverages skeleton attention to refine channels and spatial features and considers the balance between learning capability and parameter quantity. InfoGCN \cite{chi2022infogcn} emphasizes the importance of paying attention to the intrinsic connection of joints, indicating that GC using only external topology will lead to serious inefficiency and information loss in message transmission. Nevertheless, spatial-temporal information on joints in different frames is not yet considered in the above work. FD-GCN \cite{gao2021skeleton} offers a Focusing-Diffusion Graph Convolutional structure to achieve spatial-temporal attention, but the approach ignores differences among channels, which causes information redundancy. A spatial-temporal attention module is proposed in EfficientGCN \cite{Effi} to represent action-specific correlations with fewer parameters. However, compared with CTR-GCN or InfoGCN, the limited performance and scalability of channel-fixed attention weaken the discriminative power in the channel view. 

The 3D CNN-based methods are also a hot spot in the skeleton-based action recognition task. PoseC3D \cite{duan2022revisiting} uses the 3D CNN instead of graph convolution, but the heatmap of the input takes up more memory. Besides, compared with the newer GCN method, the 3D CNN-based methods have not achieved a significant performance advantage but occupy more parameters \cite{CTRGCN}. Therefore, the GCN-based methods are still competitive when the above problems are solved.

\subsection{Loss functions for fine-grained tasks} 

\begin{figure}[t]
\centering
\includegraphics[width=1\columnwidth]{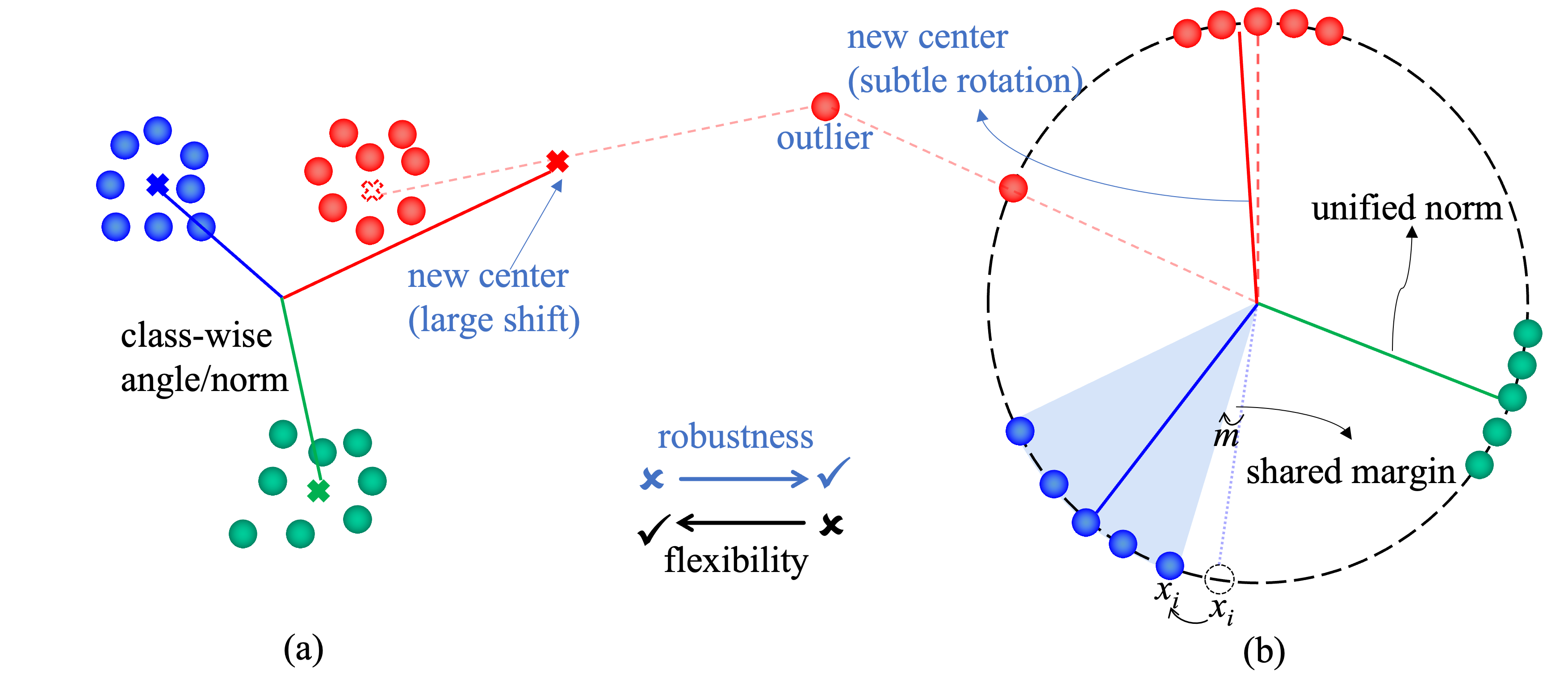} % Reduce the figure size so that it is slightly narrower than the column. Don't use precise values for figure width. This setup will avoid overfull boxes.
\caption{Schematic diagram of sample spatial distribution with (a) explicit and (b) implicit loss. The thin red dotted line illustrates the effects of the outlier. The black and blue colors demonstrate the problem of flexibility and robustness, respectively.}
\label{fig1}
\end{figure}

In recent years, researchers have realized that the softmax loss lacks the capability of learning discriminative embedding for fine-grained classification \cite{masi2018deep, wang2021deep}. Existing embedding losses fall into two major categories: explicit and implicit ones. Figure \ref{fig1} shows the characteristics of the two kinds of losses. (a) Early explicit methods \cite{hadsell2006dimensionality, sohn2016improved, hermans2017defense, wang2017deep} commonly employ siamese networks with lots of parameters to get faithful discriminative embedding. To fully utilize class annotation, some later supervised embedding losses are proposed by attaching to a classification loss. As a classical method, Center loss \cite{wen2016discriminative} is frequently used for fine-grained classification by enhancing the discriminant power in the loss layer of networks. Nevertheless, the discriminative embedding in the Euclidean space of the above approaches always suffers from the value imbalance between the center loss version and cross-entropy-based loss, in addition to weak robustness by outliers or complex feature distributions. (b) To relieve the above issues, the angular-margin-based approach \cite{liu2016large, wang2017normface, liu2017sphereface, wang2018additive, deng2019arcface} may be the better choice, which aims to expand the angle interval of classes to realize the optimization. Sphereface \cite{liu2017sphereface} leverages multiplying marginal parameters by the intra-class angle to realize the implicit embedding loss. For feasible optimization, the following methods \cite{wang2018cosface, deng2019arcface} involve an additional angular margin to accelerate the training process. However, the empirically scalable and marginal hyperparameters limit the discriminant capability of implicit loss. Recent methods \cite{zhang2019adacos, liu2019fair, peeples2022learnable} employ adaptive hyperparameters to improve the robustness. However, all these implicit methods have not noticed the class-wise discriminant on both angular and norm-based aspects, especially for large intra-class and slight inter-class variance of samples. Compared with (a), our method measures the features through the angle and norm scale, so we have more adjustment space to achieve the same optimization goal as (a) while reducing the deviation of the center's abnormal point and maintaining the advantage of flexibility.

\section{Method}

\begin{figure*}[t]
\centering
\includegraphics[width=0.9\textwidth]{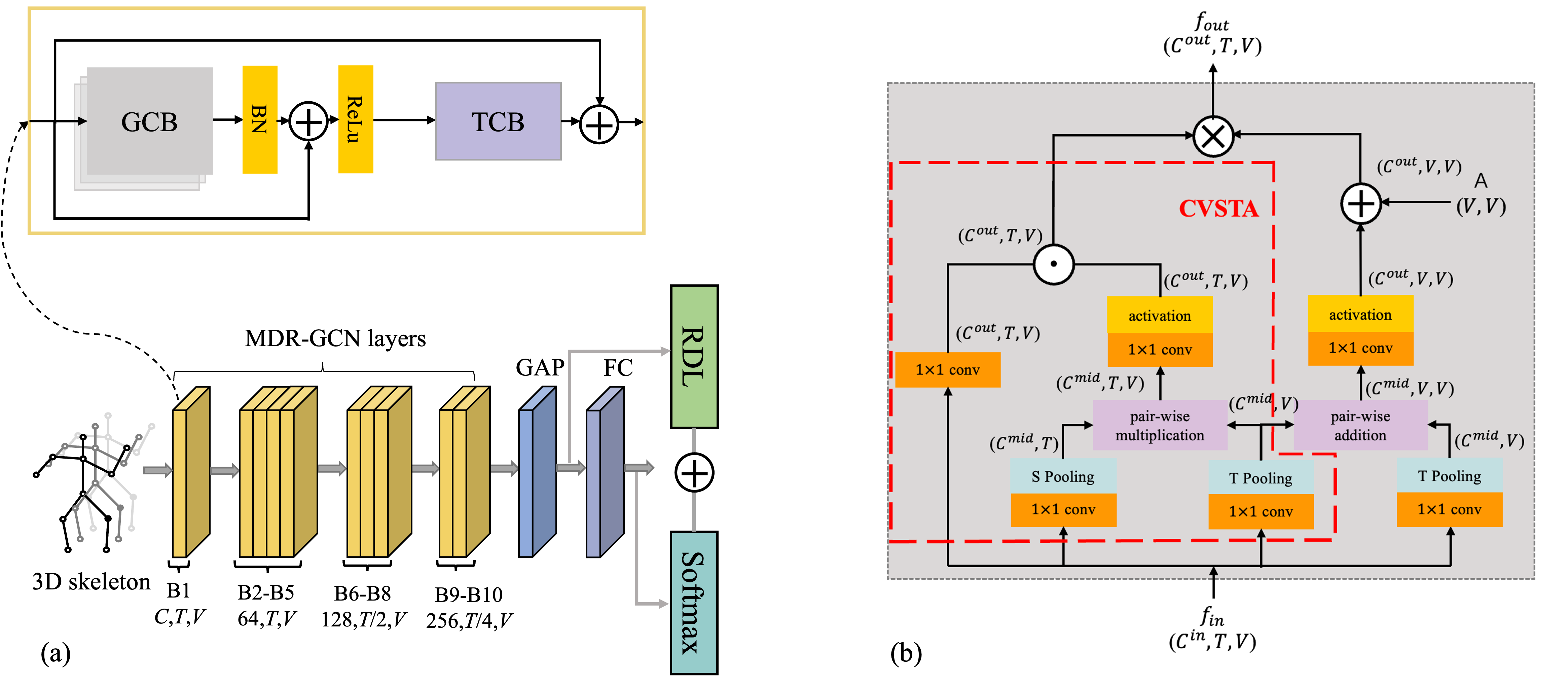} % Reduce the figure size so that it is slightly narrower than the column.
\caption{(a) Illustration of our method. The graph of skeleton sequences is fed into 10 MDR-GCN layers (B1-B10). GAP represents the global average pooling layer. GCB illustrates the graph convolution block, which consists of the summation of convolution operations on three subsets of learnable topology. TCB indicates the temporal convolution block. The proposed RDL is used in parallel with softmax. (b) The framework of our graph convolution block (GCB) in one subset of learnable topology. The proposed CVSTA is in the red dotted box. S Pooling and T Pooling denote average pooling over spatial and temporal dimensions.}
\label{fig2}
\end{figure*}

In this section, we first formulate conventional graph convolution. Then, we elaborate Channel-Variable Spatial-Temporal Attention (CVSTA) and Multi-Dimensional Refinement Graph Convolution Network (MDR-GCN). Finally, we introduce Robust Decouple Loss (RDL). The overall architecture of our model is shown in Figure \ref{fig2}.

\subsection{Preliminaries}

The skeletal graph $\mathcal G=\left(\mathcal V,\mathcal E\right)$ is established according to the natural connections of the human skeleton, where ${\mathcal V} \in {\{v_1,v_2,...,v_n\}}$ denotes the set of $V$ skeletal nodes. Set of edges $\mathcal E$ is formulated as a corresponding adjacency matrix ${\rm A}\in {\mathbb{R}^{V\times V}}$ with the element $a_{ij}$ which reflects the degree of relevance between $v_i$ and $v_j$ ($i,j \in\{1,2,...,V\}$). In spatial view, the graph convolution operation on node $v_i$ is expressed as

\begin{equation}
f_{out}(v_i)=\sum\limits_ {j = 1}^{V} {a_{ij}f_{in}(v_j)\rm{W} },
\label{fout}
\end{equation}

where $f_{in}(v_j)\in {\mathbb{R}^{C}}$ denotes the input features of $v_j$. ${\rm{W}} \in {\mathbb{R}^{C\times C'}}$ represents the weight vector of the $1 \times 1$ convolution operation, which transfers the number of input features from $C$ to $C'$.

\subsection{Model implementation}

The above spatial graph convolution process is not intuitive for task-adapted GCN of action analysis which is introduced as follows. Concretely, a $T$ frames input sample ${\rm f}_{in}$ with $C^{in}$ channels is a $C^{in}\times T\times V$ tensor. Spatial-temporal graph convolution can be defined by transformed Eq.\ref{fout} as

\begin{equation}
{\rm f}_{out}=\sum\limits_ {k=1}^{K_v}{\rm W}_k{\rm f}_{in}{\rm A}_k,
\label{fout2}
\end{equation}

where ${\rm f}_{out}\in \mathbb{R}^{C^{out}\times T\times V}$ is the output feature tensor with $C^{out}$ channels. ${\rm W}_k\in \mathbb{R}^{C^{out}\times C^{in}}$ denotes the weight matrix to adjust the number of learnable topology subsets $K_v$, where $k\in {\{1,2,..., K_v\}}$. Our ${\rm A}_k$ is initialized the same as ST-GCN \cite{STGCN}.

\textbf{Channel-Variable Spatial-Temporal Attention} To avoid limiting the channel power for the fine-grained action recognition task, our CVSTA reinforces attention to extract more discriminative spatial-temporal features of the input ${\rm f}_{in}$. CVSTA consists of three parts: discriminative spatial-temporal saliency representation, feature transformation, and feature modeling. The implementation details are described as follows.

\emph{Discriminative Spatial Temporal Saliency Representation.} To obtain discriminative representation, we first involve $1\times 1$ convolution to generate variable $C^{mid}$ channels of current middle layer for distinctive layer and its ${\rm f}_{in}$ (This means a different layer and its ${\rm f}_{in}$ may set a different $C^{mid}$). We claim that the middle layer is necessary for enhancing the discriminative representation, and the value of $C^{mid}$ may always be different from that of $ C^{out}$. This is because spatial-temporal saliency representation, refinement tensor, and spatial attention should not share the same $ C^{out}$ for their maximum discriminative channel power in a model.

Then, feature compaction is achieved by generating the pooled temporal vector ${\rm f}_T\in \mathbb{R}^{C^{mid}\times T\times 1}$ and the spatial one ${\rm f}_V\in \mathbb{R}^{C^{mid}\times 1\times V}$ which are calculated by average pooling along with the spatial dimension and the frame direction of ${\rm f}_{in}$. With ${\rm f}_T$ and ${\rm f}_V$, the saliency weights of spatial-temporal joints ${\rm f}_R \in \mathbb{R}^{C^{out}\times T\times V}$ can be calculated as

\begin{equation}
{\rm f}_R={\rm W}_{TV} \sigma ({\rm f}_T *{\rm f}_V),
\label{fr}
\end{equation}

where ${{\rm W}_{TV}} \in \mathbb{R}^{C^{out}\times C^{mid}}$ denotes the parameters of the $1 \times 1$ convolution. $\sigma$ is the activation function, and $*$ indicates pair-wise multiplication.

\emph{Feature Transformation.} In parallel with saliency representation, feature transformation is implemented by transforming ${\rm f}_{in}$ to the output tensor ${\rm f}_{TV}\in \mathbb{R}^{C^{out}\times T\times V}$ by $1\times 1$ convolution for computing the following refinement tensor.

\emph{Feature Modelling.} To achieve CVSTA, we finally adopt pair-wise element-level fusion of ${\rm f}_R$ and ${\rm f}_{TV}$ for obtaining spatial-temporal refinement tensor ${\rm R}\in \mathbb{R}^{C^{out}\times T\times V}$, which is expressed as  

\begin{equation}
{\rm R}={\rm f}_R\odot {\rm f}_{TV},
\label{frr}
\end{equation}

where $\odot$ represents the Hadamard product. In CVSTA, the number of input, middle, and output channels entirely relies on the deployed model, therefore, could flexibly deploy on different graph convolutional models.

\textbf{Graph Convolution Block} The method for refining the spatial topology of joints is similar to CVSTA. Like the operation of generating ${\rm f}_{V}$, we use convolution, pooling, and shape transformation operations on ${\rm f}_{in}$ to get ${\rm f}_S\in \mathbb{R}^{C^{mid}\times V\times 1}$. We represent the spatial correlations of the motions as the spatial tensor ${\rm \widetilde A} \in \mathbb{R}^{C^{out}\times V\times V}$, which is formulated as

\begin{equation}
{\rm \widetilde A}={\rm W}_S\sigma ({\rm f}_V \oplus {\rm f}_S),
\label{fA}
\end{equation}

where ${\rm W}_S \in \mathbb{R}^{C^{out}\times C^{mid}}$ denotes the parameters of convolution. $\oplus$ represents the pair-wise addition. Note that we share ${\rm f}_V$ of CVSTA as a spatial pooled vector which could reduce the cost of computation and offers an additional spatial-temporal relationship for ${\rm f}_S$ and ${\rm f}_T$. Combining CVSTA with spatial attention, our graph convolution operation can be written as

\begin{equation}
{\rm f}_{out}=\sum\limits_ {k=1}^{K_v}{\rm R}_k({\rm A}_k+\alpha_k {{\rm \widetilde A}_k}),
\label{fout3}
\end{equation}

where ${\rm R}_k$ and ${{\rm \widetilde A}_k}$ indicate ${\rm R}$ and ${\rm \widetilde A}$ of the $k^{th} $ subset ($ k \in {\{1,2,...,K_v\}}$), respectively. $\alpha_k$ is a learnable parameter that could balance the value between ${{\rm \widetilde A}_k}$ and shared ${\rm A}_k$. Our GCB is achieved by the above channel, spatial-temporal, and spatial refinements.

\textbf{Multi-Dimensional Refinement Graph Convolution Network} A complete MDR-GCN layer comprises a graph convolutional block (GCB), a temporal convolutional block (TCB), and a residual structure. To enhance the learning of the relationships between nearby frames, there are $m$ kernels in all temporal convolutions (kernel size $\in \{3,5,...,2m+1\}$, $m=2$ in our work).

In MDR-GCN, there are a total of 10 MDR-GCN layers. The numbers of output channels for each layer are ordered by 64-64-64-64-128-128-128-256-256-256. A global average pooling layer and a softmax classifier combing with RDL embedding loss are performed at the end.

\subsection{Robust Decouple Loss}

To further enhance the discriminative capability of our fine-grained classification network, we propose the Robust Decouple Loss, which combines both numerical-balanced angular and norm-based losses replacing the center loss. Angle losses utilize both intra-class angle loss $\mathcal{L}_{A_{in}}$ and inter-class one $\mathcal{L}_{A_{out}}$ to optimize margins among classes in the angle view. Along with the norm aspect, $\mathcal {L}_{l}$ is designed for coinciding intra-class-sample norms. 

\begin{equation}
\mathcal{L}_{A_{in}}=\frac{1}{N}\sum\limits^{N}_{i=1}(1-cos\left\langle x_i,c_{y_i}\right\rangle)^2,
\label{lain}
\end{equation}

where $N$ is the size of a mini-batch, and $x_i\in \mathbb{R}^D$ indicates the final output of the FC layer with $D$ features of the $i^{th}$ sample with the corresponding label $y_i$ for $M$ classes, $y_i \in \{1, 2,...,M\}$. $c_{y_i}\in \mathbb{R}^{D}$, which represents the center of the ${y_{i}}^{th}$ class, is randomly initialized and updated in the training process.

In addition, inter-class angular loss $\mathcal{L}_{A_{out}}$, which is involved in enlarging the margins among different classes, can be written as

\begin{equation}
\mathcal{L}_{A_{out}}=-\frac{1}{N}\sum\limits^{N}_{i=1}(1-\frac{1}{M-1}\sum\limits^{}_{k\not ={i}}cos\left\langle x_i,c_{k}\right\rangle).
\label{laout}
\end{equation}

Compared with the penalized square form of $\mathcal{L}_{A_{in}}$, $\mathcal{L}_{A_{out}}$ is free of the square to equalize the inter-class angular margins in the training process.

For one FC feature $x$, the boundaries of $\mathcal{L}_{A_{in}}$ and $\mathcal{L}_{A_{out}}$ are (0, 4) and (0, 2), respectively. A large $D$ ($D=256$ in our work) always causes large $||x||$, $||c||$ and $||x||-||c||$. Thus, $||x||-||c||$ is not suitable for numerical balance because of its large value of the initial weights. Fractional formed loss between $||x||$ and $||c||$ is more reasonable in numerical consideration. The intra-class norm-specific loss $\mathcal{L}_{l}$ can be expressed as

\begin{equation}
\beta_i =\frac{||x_i||}{||c_{y_i}||+\varepsilon}
\label{betai}
\end{equation}

\begin{equation}
\mathcal{L}_{l}=\frac{1}{N}\sum\limits^{N}_{i=1}(1-\beta_i)^2,
\label{ll}
\end{equation}

where $\beta_i$ denotes the robust ratio expression. This fraction design eliminates the weak robustness in orders of magnitude between the FC feature and its center. $\beta_i \rightarrow 1$ indicates similar norms of the two elements in $\beta_i$. A small parameter $\varepsilon$ is utilized to avoid division by zero. Nevertheless, the value of $||c_{y_i}||$ is always large enough to set $\varepsilon=0$ in most cases.

By combining $\mathcal{L}_{A_{in}}$, $\mathcal{L}_{A_{out}}$ and $\mathcal{L}_{l}$, RDL implements robust class-wise discriminative embedding for strengthening the fine-grained classification. RDL is computed as

\begin{equation}
\mathcal{L}_{R}=\mathcal{L}_{A_{in}}+{\lambda_1}\mathcal{L}_{A_{out}}+{\lambda_2}\mathcal{L}_{l}.
\label{lr}
\end{equation}

Motivated by fine-grained angular loss, RDL is dominated by $\mathcal{L}_{A_{in}}$ appending auxiliary $\mathcal{L}_{A_{out}}$ and $\mathcal{L}_{l}$. The hyperparameters ${\lambda_1}<1$ and ${\lambda_2}<1$ (always setting ${\lambda_1}={\lambda_2}=\frac{1}{N}$) are used to weigh the three losses. Inter-class norm-based loss, which leads to a numerical imbalance issue, is free of design in this paper. Consequently, the fine-grained action classification loss $\mathcal{L}$ can be achieved by combining softmax loss $\mathcal{L}_S$ and embedding loss RDL. 

\begin{equation}
\mathcal{L}_S=-\sum\limits^{N}_{i=1}log\frac{e^{{\widetilde W}_{{y_i}}^{T}x_i+b_{y_i}}}{\sum^{M}_{j=1}e^{\widetilde W_{j}^Tx_i+b_j}}
\label{ls}
\end{equation}

\begin{equation}
\mathcal{L}=\mathcal{L}_S+\mathcal{L}_{R},
\label{lfunc}
\end{equation}

where $\widetilde W_{j} \in \mathbb{R}^D$ denotes the $j^{th}$ column of the weights $\widetilde W \in \mathbb{R}^{D\times M}$ in the last FC layer and $b\in \mathbb{R}^M$ is the bias term. Finally, the loss function $\mathcal{L}$ can achieve class-wise discriminative learning for fine-grained classification.

\subsection{Gradients calculation of RDL}

In this subsection, we describe the derivation process of RDL. According to Eq.\ref{lr}, the gradients of RDL are the sum of three items, which can be expressed as

\begin{equation}
\frac{\partial \mathcal{L}_{R}}{\partial x_i}=\frac{\partial \mathcal{L}_{A_{in}}}{\partial x_i}+{\lambda_1}\frac{\partial \mathcal{L}_{A_{out}}}{\partial x_i}+{\lambda_2}\frac{\partial \mathcal{L}_{l}}{\partial x_i}.
\label{gradRx}
\end{equation}

For $c_{y_i}$, as $\mathcal{L}_{A_{in}}$ and  $\mathcal{L}_{A_{out}}$ with respect to $c_{y_i}$ is symmetrical to $x_i$, the gradients in the angle items with respect to $c_{y_i}$ is the similar to $x_i$. Therefore, we only need to pay extra attention to the update equation of $c_{y_i}$ in $\mathcal{L}_{l}$. Then, we solve for the gradients of each item.

\textbf{The gradients of $\mathcal{L}_{A_{in}}$} $cos\left\langle x_i,c_{y_i}\right\rangle$ in $\mathcal{L}_{A_{in}}$ can be transfered as

\begin{equation}
cos\left\langle x_i,c_{y_i}\right\rangle=\frac{x_i^Tc_{y_i}}{||x_i||||c_{y_i}||}.
\label{cosf}
\end{equation}

Based on Eq.\ref{lain}, $\mathcal{L}_{A_{in}}$ can be expressed as

\begin{equation}
\mathcal{L}_{A_{in}}=\frac{1}{N}\sum\limits^{N}_{i=1}(1-\frac{x_i^Tc_{y_i}}{||x_i||||c_{y_i}||})^2.
\label{lain2}
\end{equation}

The gradients of $\mathcal{L}_{A_{in}}$ with respect to $x_i$ are formulated as

\begin{equation}
\frac{\partial \mathcal{L}_{A_{in}}}{\partial x_i}=\frac{2}{N}\sum\limits^{N}_{i=1}(1-\frac{x_i^Tc_{y_i}}{||x_i||||c_{y_i}||})(\frac{{x_i}{x_i^T}}{||x_i||^3_2}-\frac{I}{||x_i||})\frac{c_{y_i}}{||c_{y_i}||},
\label{laingrad}
\end{equation}

where $I \in \mathbb{R}^{D\times D}$ indicates an identity matrix. 

\textbf{The gradients of $\mathcal{L}_{A_{out}}$}  According to Eq.\ref{laout}, $\mathcal{L}_{A_{out}}$ can be written as

\begin{equation}
\mathcal{L}_{A_{out}}=-\frac{1}{N}\sum\limits^{N}_{i=1}(1-\frac{1}{M-1}\sum\limits^{}_{k\not ={i}}(1-\frac{x_i^Tc_{y_k}}{||x_i||||c_{y_k}||})).
\label{laout2}
\end{equation}

The gradients of $\mathcal{L}_{A_{out}}$ with respect to $x_i$ are formulated as

\begin{equation}
\frac{\partial \mathcal{L}_{A_{out}}}{\partial x_i}=\frac{1}{N(M-1)}\sum\limits^{N}_{i=1}\sum\limits^{}_{k\not ={i}}(\frac{{x_i}{x_i^T}}{||x_i||^3_2}-\frac{I}{||x_i||})\frac{c_{y_k}}{||c_{y_k}||}.
\label{laoutgrad}
\end{equation}

\textbf{The gradients of $\mathcal{L}_{l}$} 

$\mathcal{L}_{l}$ is expressed as

\begin{equation}
\mathcal{L}_{l}=\frac{1}{N}\sum\limits^{N}_{i=1}(1-\frac{||x_i||}{||c_{y_i}||+\varepsilon})^2.
\label{ll2}
\end{equation}

The gradients of $\mathcal{L}_{l}$ concerning $x_i$ are formulated as

\begin{equation}
\frac{\partial \mathcal{L}_{l}}{\partial x_i}=\frac{2}{N}\sum\limits^{N}_{i=1}(\frac{||x_i||}{||c_{y_i}||+\varepsilon}-1)\frac{x_i}{||x_i||(||c_{y_i}||+\varepsilon)}.
\label{llxi}
\end{equation}

And the gradients of $\mathcal{L}_{l}$ with respect to $c_{y_i}$ are represented as

\begin{equation}
\frac{\partial \mathcal{L}_{l}}{\partial c_{y_i}}=\frac{2}{N}\sum\limits^{N}_{i=1}(1-\frac{||x_i||}{||c_{y_i}||+\varepsilon})\frac{||x_i||c_{y_i}}{||c_{y_i}||(||c_{y_i}||+\varepsilon)^{2}}.
\label{llci}
\end{equation}

\section{Experiments}

\subsection{Datasets}

\textbf{FineGym99} FineGym99 \cite{shao2020finegym} is a fine-grained action recognition dataset containing 29k videos of 99 fine-grained gymnastics action categories. Compared to existing action recognition datasets, it provides temporal annotations at both action and sub-action levels with a three-level semantic hierarchy. We conduct experiments using 3D-pose extracted the same as Pyskl \cite{duan2022pyskl}. The dataset is publicly available at https://sdolivia.github.io/FineGym/.

\textbf{FSD-10} To fully evaluate the effectiveness of spatial-temporal modules in our network, FSD-10 \cite{liu2020fsd} is involved in our experiments. FSD-10 collects 1484 clips from the worldwide figure skating championships in 2017–2018 and contains ten fine-grained actions in men/ladies’ programs. Each clip is at 30 fps with a resolution of $1080\times 720$. There are 1500 frames for each sample and 25 joints for each frame. This dataset has a significant duration variance. Therefore we can better verify the performance of our method in the temporal view. The dataset is publicly available at https://shenglanliu.github.io/fsd10/.
I
\textbf{NTU RGB+D} NTU RGB+D \cite{shahroudy2016ntu} is a large-scale human action recognition dataset that contains 56800 clips of actions. The action samples are categorized into 60 classes, where 50 classes have single-person actions, and the rest are pair interactive actions. We adopt 25 joints to represent each frame of one person (no more than two persons). We recommend two split versions for conducting experiments, i.e., cross-subject and cross-view. In cross-subject, the dataset is of 40320 training samples and 16560 testing samples. The training set of X-view obtains 37920 samples via the $2^{nd}$ and $3^{rd}$ cameras. The $1^{st}$ camera generates the corresponding testing set, which consists of 18960 samples. We utilize both two versions for experiments in this paper. The dataset is publicly available at https://rose1.ntu.edu.sg/dataset/actionRecognition/.

\begin{table}[b]
\caption{Comparisons of accuracies when adding CVSTA or RDL from ST-GCN and CTR-GCN.}
\centering
%\resizebox{.95\columnwidth}{!}{
\begin{tabular}{cccc}
   \toprule
   Models & CVSTA & RDL & Accuracy(\%)  \\
   \midrule
   \multirow{4}*{ST-GCN} & $\times $ & $\times $ & 86.59  \\
   ~ & \checkmark & $\times $ & 88.71  \\
   ~ & $\times $ & \checkmark & 87.29  \\
   ~ & \checkmark & \checkmark & 89.65  \\
     \midrule
   \multirow{4}*{CTR-GCN} & $\times $ & $\times $ & 90.59  \\
   ~ & \checkmark & $\times $ & 90.82  \\
   ~ & $\times $ & \checkmark & 92.00  \\
   ~ & \checkmark & \checkmark & 92.94  \\
   \bottomrule
\end{tabular}
\label{tab1}
\end{table}

\begin{table}[t]
\caption{Comparisons of the validation accuracy when utilizing CVSTA with different $C^{mid}$, $C^{mid}=fixed$ indicates raising the number of input channels directly to the output. }
\centering
%\resizebox{.95\columnwidth}{!}{
\begin{tabular}{ccc}
   \toprule
    Settings & $C^{mid}$ & Accuracy(\%) \\
    \midrule
    Cf & $fixed$ & 90.59 \\
    C16 & $C^{in}/16$ & 92.71 \\
    C8 & $C^{in}/8$ & \textbf{93.88} \\
    C4 & $C^{in}/4$ & \textbf{93.88} \\
      \bottomrule
\end{tabular}

\label{tabref2}
\end{table}

\begin{table}[t]
\caption{Performance of utilizing difference attention block on our framework.}
\centering
%\resizebox{.95\columnwidth}{!}{
\begin{tabular}{cccc}
   \toprule
    Settings & Attentions & Models & Accuracy(\%) \\
    \midrule
    A & SA-GC &  InfoGCN \cite{chi2022infogcn} &91.9\\
    B & ST-JointAtt &  Efficient-GCN b4 \cite{Effi} & 92.6\\
    C & CVTSA & MDR-GCN  & \textbf{93.3}\\
      \bottomrule
\end{tabular}
\label{tabatt}
\end{table}

\begin{table}[t]
\caption{Performance of utilizing different methods on the FSD-10 dataset with noisy data ('wo RDL' represents that setting MDR-GCN with Softmax loss only).}
\centering
%\resizebox{.95\columnwidth}{!}{
\begin{tabular}{cccc}
   \toprule
    Methods & 0\% noise (\%) & 1\% noise (\%) & 10\% noise (\%) \\
    \midrule
   CTR-GCN & 90.58 & 89.65 & 85.88 \\
    MDR-GCN (wo RDL) & 91.06 & 90.82 & 87.13\\
   MDR-GCN & \textbf{93.18} & \textbf{92.47} & \textbf{91.29}\\
      \bottomrule
\end{tabular}
\label{tabnoise}
\end{table}

\begin{table}[t]
\caption{Comparisons of accuracies when utilizing different components of RDL on the FSD-10 dataset.}
\centering
%\resizebox{.95\columnwidth}{!}{
\begin{tabular}{cccc}
   \toprule
    $\mathcal{L}_{A_{in}} $ & $\mathcal{L}_{A_{out}} $ & $\mathcal{L}_{A_l} $ & Accuracy(\%) \\
    \midrule
    $\times $ & $\times $ & $\times $ & 91.06 \\
    \checkmark & $\times $ & $\times $ & 91.76 \\
    $\times $ & \checkmark & $\times $ & 91.29 \\
    $\times $ & $\times $ & \checkmark & 90.35 \\
    \checkmark & \checkmark & $\times $ & 92.00 \\
    \checkmark & $\times $ & \checkmark & 92.94 \\
    $\times $ & \checkmark & \checkmark & 92.25 \\
    \checkmark & \checkmark & \checkmark & \textbf{93.18} \\
      \bottomrule
\end{tabular}
\label{tab3}
\end{table}

\begin{table}[t]
\caption{Comparisons of accuracies when utilizing MDR-GCN with different settings.}
\centering
%\resizebox{.95\columnwidth}{!}{
\begin{tabular}{cccc}
   \toprule
   Settings & $\sigma$ & $k$ & Accuracy(\%)  \\
   \midrule
   H1 & hardswish & 1 & 90.35  \\
   T1 & tanh & 1 & 90.82  \\
   S1 & sigmoid & 1 & 90.82  \\
   S2 & sigmoid & 2 & 90.82  \\
   T2& tanh & 2 & \textbf{91.06}  \\
   T3 & tanh & 3 & 90.59  \\
   \bottomrule
\end{tabular}
\label{tabA2}
\end{table}

\begin{table}[t]
\caption{Comparison of the efficiency of mainstream methods.}
\centering
%\resizebox{.95\columnwidth}{!}{
\begin{tabular}{cccc}
 \toprule
 Models & Params & FLOPs & Accuracy(\%) \\
 \midrule
 MS-G3D & 2.8M & 24.7G & 92.0 \\
 CTR-GCN & 1.2M & 14.4G & 91.9 \\
 PoseC3D & 2.0M & 15.9G & 93.2 \\
 MDR-GCN & 1.3M & 15.3G & 93.3 \\
  \bottomrule
\end{tabular}
\label{tabPose}
\end{table}

\begin{table}[t]
\caption{Classification accuracy comparison against other competitive loss functions on the FSD-10 and FineGym99 datasets.}
\centering
%\resizebox{.95\columnwidth}{!}{
\begin{tabular}{ccc}
   \toprule
   Methods & FSD-10(\%) & FineGym99(\%)\\
   \midrule
   Baseline & 91.06 & 90.13\\
   center loss \cite{wen2016discriminative} & 91.06 & 91.17\\
   sphereFace \cite{liu2017sphereface} & 88.24 & --\\
   LMCL\cite{wang2018cosface} & 92.24 & --\\ 
   arcFace \cite{deng2019arcface} & 92.71 & 91.52\\
   LACE \cite{peeples2022learnable} & 91.29 & 91.74\\
   \midrule
   RDL without $L_{out}$ & 92.94 & 91.94  \\
   RDL & \textbf{93.18} & \textbf{92.11}\\
   \bottomrule
\end{tabular}

\label{tab4}
\end{table}

\begin{figure}[t]
\centering
\includegraphics[width=0.9\columnwidth]{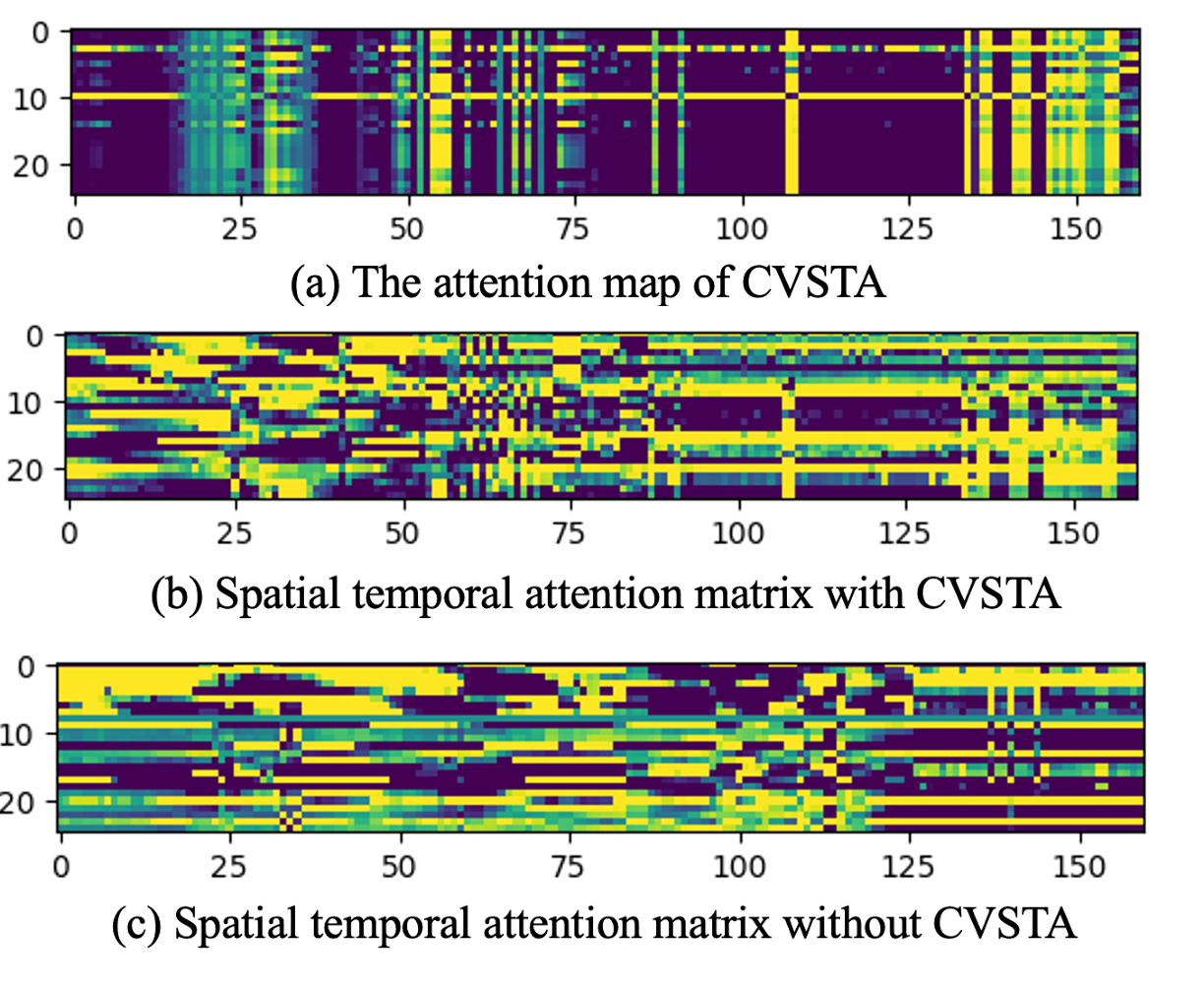} % Reduce the figure size so that it is slightly narrower than the column. Don't use precise values for figure width. This setup will avoid overfull boxes.
\caption{Examples of the attention map and spatial-temporal matrices. The warm colors indicate more intense features.}
\label{fig3}
\end{figure}

\begin{figure}[t]
\centering
\includegraphics[width=0.9\columnwidth]{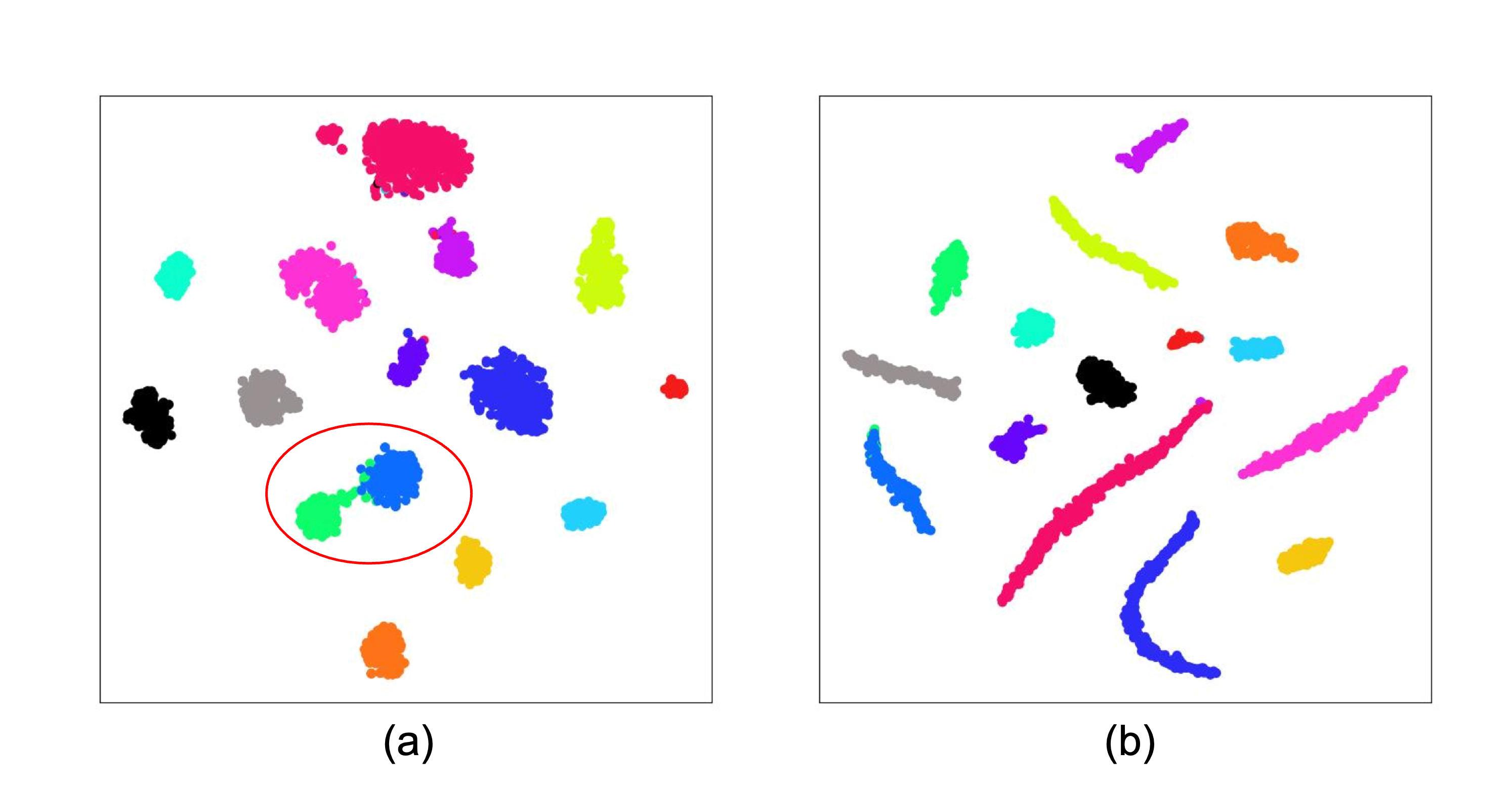} % Reduce the figure size so that it is slightly narrower than the column. Don't use precise values for figure width. This setup will avoid overfull boxes.
\caption{Feature visualization of (a) MDR-GCN with Softmax loss and (b) with RDL by t-SNE.}
\label{figtsne}
\end{figure}

\begin{table}[t]
\caption{Classification accuracy comparison against existing methods on the FineGym99 dataset.}
\centering
%\resizebox{.95\columnwidth}{!}{
\begin{tabular}{cc}
   \toprule
   Methods & Accuracy(\%)\\
   \midrule
   ST-GCN \cite{STGCN} & 36.4\\
   MS-G3D \cite{MSG3D} & 92.0\\
   CTR-GCN \cite{CTRGCN} & 91.9\\
   MS-G3D++ \cite{duan2022revisiting} & 92.6\\
   InfoGCN \cite{chi2022infogcn}& 92.0\\
   PoseC3D \cite{duan2022revisiting} &94.3 \\
   \textbf{MDR-GCN} & \textbf{94.5}\\
   \bottomrule
\end{tabular}

\label{tab6}
\end{table}

\begin{table}[t]
\caption{Classification accuracy comparison against existing methods on the FSD-10 dataset.}
\centering
%\resizebox{.95\columnwidth}{!}{
\begin{tabular}{cc}
   \toprule
   Methods & Accuracy(\%)\\
   \midrule
   ST-GCN \cite{STGCN} & 84.24\\
   2S-AGCN \cite{2SAGCN} & 88.23\\
   AS-GCN \cite{li2019actional} & 86.82\\
   MS-G3D \cite{MSG3D} & 88.72\\
   CTR-GCN \cite{CTRGCN} & 90.58\\
   InfoGCN \cite{chi2022infogcn} & 91.76\\
   \textbf{MDR-GCN} & \textbf{94.18}\\
   \bottomrule
\end{tabular}
\label{tab7}
\end{table}

\begin{table}[t]
\caption{Classification accuracy comparison against existing methods on the NTU RGB+D dataset.}
\centering
%\resizebox{.95\columnwidth}{!}{
\begin{tabular}{ccc}
   \toprule
   Methods & X-sub(\%) & X-view(\%) \\
   \midrule
   ST-GCN \cite{STGCN} & 81.5 & 88.3\\
   2S-AGCN \cite{2SAGCN} & 88.5 & 95.1\\
   AS-GCN \cite{li2019actional} & 86.8 & 94.2\\
   Shift-GCN \cite{cheng2020skeleton} & 90.7 & 96.5\\
   DC-GCN+ADG \cite{cheng2020decoupling} & 90.8 & 96.6\\
   Dynamic GCN \cite{ye2020dynamic} & 91.5 & 96.0\\
   MSIN \cite{cheng2020decoupling} & 91.5 & 96.5\\
   MS-G3D \cite{MSG3D} & 91.5 & 96.2\\
   MSG3D++ \cite{duan2022revisiting} & 92.2 & 96.6\\
   MST-GCN \cite{chen2021multi} & 91.5 & 96.6\\
   CTR-GCN \cite{CTRGCN} & 92.4 & 96.8\\
   Efficient-GCN B4 \cite{Effi} & 92.1 & 96.1\\
   InfoGCN \cite{chi2022infogcn} & 93.0 & 97.1 \\
   PoseC3D \cite{duan2022revisiting} & \textbf{94.1} & 97.1 \\
   \textbf{MDR-GCN} & 92.8 & \textbf{97.2}\\
   \bottomrule
\end{tabular}
\label{tab8}
\end{table}

\subsection{Training details}

We conducted all experiments on the PyTorch deep learning framework. Stochastic gradient descent (SGD) with Nesterov momentum (0.9) is applied as the optimization strategy, and the learning rate is set to 0.1. For FineGym99, we set the batch size to 64, and the learning rate is divided by ten at 60 and 120 epochs. For FSD-10, the learning rate multiply by 0.1 at epochs 150 and 225 for 300 epochs, 256 non-zero frames, and 48 batch size is set in Sec 4.3, Sec 4.4, all frames and 8 batch size are utilized to show the complete performance in Sec 4.5, Sec 4.6. For NTU RGB+D, the batch size is set to 64, and the learning rate decays with a factor of 0.1 at epochs 35 and 55 for a total of 65 epochs.

\subsection{Ablation study}

\textbf{Expandability of our method} To evaluate the effectiveness of the CVSTA block, we plugged it into ST-GCN \cite{STGCN} and CTR-GCN \cite{CTRGCN}. As can be seen in Table \ref{tab1}, the experimental results of plugged ST-GCN and CTR-GCN can improve performance than their corresponding original versions on the FSD-10 dataset. Compared with the original ST-GCN, CVSTA plus ST-GCN can reach an enormous improvement of $2.12\%$. CTR-GCN can also be enhanced performance ($+0.23\%$) by combining with CVSTA. The above experimental results demonstrate that CVSTA is effective for fine-grained action recognition, especially for the attention-free method (e.g., ST-GCN). For evaluating the effectiveness of the proposed discriminative loss, RDL is attached to ST-GCN, CTR-GCN, and CTR-GCN+CVSTA. As shown in Table \ref{tab1}, the three RDL-attached methods exceed corresponding ST-GCN, CTR-GCN, and CTR-GCN+CVSTA by $0.7\%$, $1.42\%$ and $2.12\%$, respectively. According to the above experimental results, CVSTA and RDL plugged methods can outperform the baseline models. This illustrates that CVSTA can increase spatial-temporal discrimination. Besides, RDL can achieve more excellent performance by combining the discrimination-reinforced networks. Thus, CVSTA is beneficial for reinforcing the power of RDL for our fine-grained task.

\textbf{Effectiveness of the variable channel} We utilize different $C^{mid}$ in CVSTA with RDL to examine the effect of changing channels. As shown in Table \ref{tabref2}, (1) compared to the fixed setting (setting Cf), any configuration which utilizes a variable channel outperforms the fixed setting, indicating the variable channels enhance the capability of perceiving the latent and filtering out the redundant features. (2) comparing settings C16 and C8, we find that compression of the channel results in abandoning valid information. Therefore, we choose C8 collocation as our configuration, considering both performance and space consumption.

\textbf{Effectiveness of our spatial-temporal attention} To ascertain that CVSTA is providing more high-quality features, we train our framework by using different attention on the FineGym99 dataset. As Table \ref{tabatt} shows, settings B and C show higher performance compare to the non-spatial-temporal SA-GC attention block (setting A). And comparing settings B and C, we see that CVSTA consistently outperforms the spatial-temporal attention which being utilized after GCN and TCN layers. 

\textbf{Effectiveness of our method for noisy data} To have a deeper understanding of the robustness of our model, we modify the FSD-10 dataset by setting the random coordinates in some (1\%, 10\%)  skeletal data to 0. Table \ref{tabnoise} shows the performance of our method on the modified-FSD-10 dataset. Based on the comparison between 1\% and 10\%, we can see when there are more outliers on a dataset, our method can reflect obvious performance advantages.

\textbf{Effectiveness of RDL} We set the performance of MDR-GCN without RDL ($91.06\%$) as the baseline to investigate the effectiveness of RDL terms via setting $\sigma=tanh$, $k=2$. Table \ref{tab3} illustrates angular terms of RDL can enhance the performance by comparing with the baseline. In contrast, norm-specific RDL without angular terms would reduce the performance ($-0.71\%$). This illustrates angular terms are more important than the norm-based one for the fine-grained action recognition task. Besides, the norm-based term is helpful for the angular terms of RDL ($+1.18\%$). The results and analyses coincide with our design of RDL.

\textbf{Configuration of MDR-GCN} Table \ref{tabA2} shows the effects of activation function $\sigma$ and the number of TCN blocks $k$ in different settings. As shown in Table \ref{tabA2}, various settings achieve similar results, illustrating our model's robustness. The details are as follows. According to the experimental results of H1, T1, and S1 collocations, both tanh and sigmoid activation functions are acceptable options. By adjusting $k$, the different results of both S2 and T2 indicate that the tanh activation function is superior to others. Besides, the results of variable $k$ in T1, T2, and T3 encourage us to choose multi-kernel (i.e., $k=2$) for MDR-GCN. However, a large $k$ may reduce the performance because of over-learning and enlarging the network's parameters. Considering both performance and efficiency, we choose T2 collocation as our configuration.

\textbf{Efficiency of our method} In performance comparison between our method and C3D, we adopt the input shape $48 \times 56 \times 56$ for PoseC3D, $3 \times 64 \times 17$ for MS-G3D, CTR-GCN and our method. Table \ref{tabPose} shows that under such configuration, our method achieves more competitive performance and efficiency on the FineGym99 dataset.

\subsection{Comparison with other loss functions}

Table \ref{tab4} shows the performances of advanced loss functions on FSD-10 and FineGym99. For a fair comparison, we use the C8 configuration of Table \ref{tabref2} in the rest experiments of this paper. We observe that (1) on FSD-10, RDL gains a $2.12\%$ improvement over the center loss, which indicates the effectiveness of the proposed embedding loss. Besides, our $\mathcal{L}_{A_{in}} +\mathcal{L}_{l} $ loss version (RDL without $\mathcal{L}_{A_{out}} $) and center loss contain two similar optimization properties (i.e., angle and norm). In this case, $\mathcal{L}_{A_{in}}+\mathcal{L}_{l} $ also outperforms center loss by $1.88\%$, which illustrates the robustness of our loss design. (2) RDL achieves an accuracy of $93.18\%$, which surpasses the state-of-the-art implicit losses in recent years, including competitive arcFace loss ($+0.47\%$). To further illustrate the effectiveness of RDL, we involve the competitive arcFace and recent LACE for comparison on the FineGym99 dataset. RDL outperforms arcFace and LACE by $0.59\%$ and $0.37\%$, respectively. The results of FineGym99 indicate that RDL is adequate and robust for fine-grained datasets with large-scale classes.

\subsection{Visualization of CVSTA}

We obtain the experimental sample by trimming '3StepSequence3' of FSD-10 to visualize the attention map of CVSTA (Figure \ref{fig3} (a)). Frame extraction strategy is excluded to ensure the frame coincides with the pose temporal position of the original action. We reveal the first 160 frames as the sample in the untrimmed joint sequence, which includes most of the continuous key poses. We further illustrate that the first 120 frames of the sample express a sliding sequence part, and the rest represent a complex technical sequence. The $10^{th}$ joint (right knee) should be highlighted in most frames because of the intuitive plain sequence. As shown in Figure \ref{fig3}, visualization of the attention map provides a clear focus in the $10^{th}$ joint row. The visualization results illustrate that CVSTA is effective for spatial joints.

Furthermore, a few body twists are performed in the temporal neighborhood of the $75^{th}$ frame, leading to highlighted shoulder joints by CVSTA during this period. Thus, we can confirm that CVSTA achieves spatial-temporal attention. Compared with the sliding sequence part, most of the frames of the technical sequence part are highlighted in the attention map. This shows that our CVSTA provides additional attention to the temporal dimension. As shown in Figure \ref{fig3} (b) and (c), the visualization of ${\rm f}_{TV}$ with CVSTA also shows more excellent results in the above position than those without CVSTA. This indicates CVSTA is a benefit for extracting key frames to cope with large duration variance action samples.

\subsection{Visualization of RDL}

We use the MDR-GCN model on the FineGym99 dataset, which employs (a) Softmax loss and (b) RDL as loss functions, and perform t-SNE \cite{van2008visualizing} dimensionality reduction on the generated features for 15 classes with more samples. Figure \ref{figtsne} shows the results after dimensionality reduction. We observe that (1) there are significantly more outliers in (a) than in (b), which reflects the good robustness of RDL. (2) For some hard-to-distinguish classes (such as the two classes inside the red circle in (a)), RDL can better achieve class separability.

\subsection{Comparison with the State-of-the-Art}

In this subsection, we compare our method with the state-of-the-art skeleton-based action recognition methods on all three datasets introduced above. FineGym99 and FSD-10 are utilized in fine-grained tasks to evaluate our approach's advantages on spatial-temporal joints and large-scale classes. To show the generality of our model, the most widely used dataset NTU RGB+D is employed for comparison. On NTU RGB+D and FineGym99 dataset, our performance is fused by the results of multiple skeletal modalities as the mainstream methods \cite{CTRGCN, duan2022revisiting, chi2022infogcn}.

The results are shown in Tables \ref{tab6}, \ref{tab7} and \ref{tab8}. Our model achieves state-of-the-art performance with a large margin on the Fine-Grained datasets. And on NTU RGB+D, our method also gets close to the state-of-the-art models considering both evaluation benchmarks. This illustrates that, as a fine-grained solution, our method can still preserve good capability on coarse-grained datasets.

\section{Limitations and Social Impacts}

Currently, we only explore the performance of RDL on skeleton-based action recognition, but RDL may also be applied in other fine-grained tasks, which need to be explored in future work. Furthermore, although RDL improves the robustness problem by decoupling the Euclidean distance as an explicit method, the effect of outliers on the center still exists.

Our method achieves a significant improvement in the accuracies of fine-grained action recognition, which could provide a new solution for recognizing complex and similar actions (such as technical actions in gymnastics, diving, figure skating, etc.) that are common in reality. There are no known socially detrimental effects of our work other than those typically associated with developing new AI systems.

\section{Conclusion}

This work proposes a Multi-Dimensional Refinement Graph Convolution Network for fine-grained skeleton-based action recognition (MDR-GCN), including a Channel-Variable Spatial-Temporal Attention (CVSTA). The model is powerful for extracting spatial and temporal discriminative features. Furthermore, we propose a Robust Decouple Loss, which can enhance intra-class compactness and inter-class separability for the fine-grained recognition task. Our method outperforms the existing skeleton-based approaches on the three challenging datasets. Meanwhile, the flexibility of RDL and CVSTA could improve future work.

%\section*{Acknowledgments}
%This should be a simple paragraph before the References to thank those individuals and institutions who have supported your work on this article.

{

\bibliographystyle{IEEEtran}
\bibliography{IEEEabrv, egbib}

%\newpage

 %needed around the contents of the optional argument to biography to prevent
 %the LaTeX parser from getting confused when it sees the complicated
 %$\backslash${\tt{includegraphics}} command within an optional argument. (You can create
% your own custom macro containing the $\backslash${\tt{includegraphics}} command to make things
 %simpler here.)
 

%\bf{If you include a photo:}\vspace{-33pt}
%\begin{IEEEbiography}[{\includegraphics[width=1in,height=1.25in,clip,keepaspectratio]{fig1}}]{Michael Shell}
%Use $\backslash${\tt{begin\{IEEEbiography\}}} and then for the 1st argument use $\backslash${\tt{includegraphics}} to declare and link the author photo.
%Use the author name as the 3rd argument followed by the biography text.
%\end{IEEEbiography}



%\vfill

\end{document}